\documentclass[
]{ceurart}
\usepackage{amsmath}
\usepackage[scientific-notation=true]{siunitx}
\usepackage{graphicx,subcaption} 
\usepackage[most]{tcolorbox}

\usepackage{listings}
\begin{document}

\copyrightyear{2025}
\copyrightclause{Copyright for this paper by its authors.  Use permitted under Creative Commons License Attribution 4.0  International (CC BY 4.0).}

\conference{CLEF 2025 Working Notes, 9 -- 12 September 2025, Madrid, Spain}

\title{SINAI at eRisk@CLEF 2025: Transformer-Based and Conversational Strategies for Depression Detection}

\title[mode=sub]{Notebook for the eRisk Lab at CLEF 2025}

\author{Alba Maria Marmol-Romero}[%
orcid=0000-0001-7952-4541,
email=amarmol@ujaen.es
]

\author{Manuel Garcia-Vega}[%
orcid=0000-0003-2850-4940,
email=mgarcia@ujaen.es
]

\author{Miguel Angel Garcia-Cumbreras}[%
orcid=0000-0003-1867-9587,
email=magc@ujaen.es
]

\author{Arturo Montejo-Raez}[%
orcid=0000-0002-8643-2714,
email=amontejo@ujaen.es
]

\address{Computer Science Department, SINAI, CEATIC, University of Jaen, 23071, Spain}

\begin{abstract}
This paper describes the participation of the SINAI-UJA team in the eRisk@CLEF 2025 lab. Specifically, we addressed two of the proposed tasks: (i) Task 2: Contextualized Early Detection of Depression, and (ii) Pilot Task: Conversational Depression Detection via LLMs.
Our approach for Task 2 combines an extensive preprocessing pipeline with the use of several transformer-based models, such as RoBERTa Base or MentalRoBERTA Large, to capture the contextual and sequential nature of multi-user conversations. For the Pilot Task, we designed a set of conversational strategies to interact with LLM-powered personas, focusing on maximizing information gain within a limited number of dialogue turns.
In Task 2, our system ranked 8th out of 12 participating teams based on F1 score. However, a deeper analysis revealed that our models were among the fastest in issuing early predictions, which is a critical factor in real-world deployment scenarios. This highlights the trade-off between early detection and classification accuracy, suggesting potential avenues for optimizing both jointly in future work. In the Pilot Task, we achieved 1st place out of 5 teams, obtaining the best overall performance across all evaluation metrics: DCHR, ADODL and ASHR. Our success in this task demonstrates the effectiveness of structured conversational design when combined with powerful language models, reinforcing the feasibility of deploying LLMs in sensitive mental health assessment contexts.
\end{abstract}


\maketitle

\section{Introduction}
Every day, people share a huge amount of content on social media, making these platforms a valuable resource for detecting mental health issues and risky behaviors early on. The eRisk@CLEF 2025 lab~\cite{overview2025erisk,overview2025erisk1,overview2025erisk2} is dedicated to improving computational systems that can identify mental disorders like depression as soon as possible. This year, the lab has proposed three specific tasks to further this goal:

\begin{itemize}
    \item \textbf{Task 1 - Search for symptoms of depression}. It consists of ranking sentences from a collection of user writings according to their relevance to a depression symptom. Then, the participants will have to provide rankings for the 21 symptoms of depression from the BDI Questionnaire. It is a continuation of Task 1 proposed for eRisk 2023~\cite{erisk2023} and eRisk 2024~\cite{erisk2024}.
    \item \textbf{Task 2 -  Contextualized Early Detection of Depression}. This new task focuses on detecting early signs of depression by sequentially analyzing full conversational contexts and detecting early traces of depression as soon as possible.
    
    \item \textbf{Pilot Task - Conversational Depression Detection via LLMs}. The challenge lies in determining whether the LLM persona exhibits signs of depression, accompanied by an explanation of the main symptoms that informed their decision.
\end{itemize}

\hfill \break

This work presents the participation of our research group, the SINAI\footnote{https://sinai.ujaen.es/} team, in \textit{Task 2: Contextualized Early Detection of Depression} and \textit{Pilot Task: Conversational Depression Detection via LLMs}. The rest of the paper is organized as follows: sections \ref{sec:task2} and \ref{sec:task3} describe in detail our participation in task 2 and the pilot task, respectively. Each of them is divided into subsections in which, first, we introduce what the task consists of, the data provided, and the evaluation measures used. Secondly, the system developed and the methodology used are presented. Thirdly, the experimental setup is detailed. Subsequently, the results obtained and a discussion on them are presented. Finally, Section \ref{sec:conclusions} shows the conclusions obtained after the participation in the eRisk lab and the perspectives for future work.

\section{Task 2: Contextualized Early Detection of Depression}
\label{sec:task2}
\subsection{Task description}
This task focuses on the early detection of depression by analyzing full conversational contexts, including the contributions of all users involved in a discussion. Unlike previous editions that relied on isolated user posts, this task highlights the importance of dialogue dynamics and sequential processing of messages. The challenge is structured in two phases: a training phase with individual user writings (without context), and a test phase where models must process conversations chronologically and make real-time predictions as new messages appear. Evaluation considers both accuracy and timeliness, using metrics such as ERDE (Early Risk Detection Error), flatency, and traditional scores like precision, recall, and F1. The task encourages the development of context-aware systems suited for real-world applications in mental health monitoring.

\subsection{Dataset}
For this task, we have made use of the dataset provided by the organisers, together with one extracted specifically for this task.

\subsubsection{Dataset provided}

The provided testing data consists of a dataset in which there are two types of instances: submissions and comments. Submissions represent the primary posts created by users. They are the main content entries. Comments are the responses or replies made by users to a submission or other comments, forming a hierarchical structure. Moreover, the main objective of this phase is to classify and assign a score of risk to a target subject that sometimes can be the author of the primary post but other times can only appear in some comments.

\begin{figure}[!ht]

 \begin{subfigure}{0.49\textwidth}
     \includegraphics[width=\textwidth]{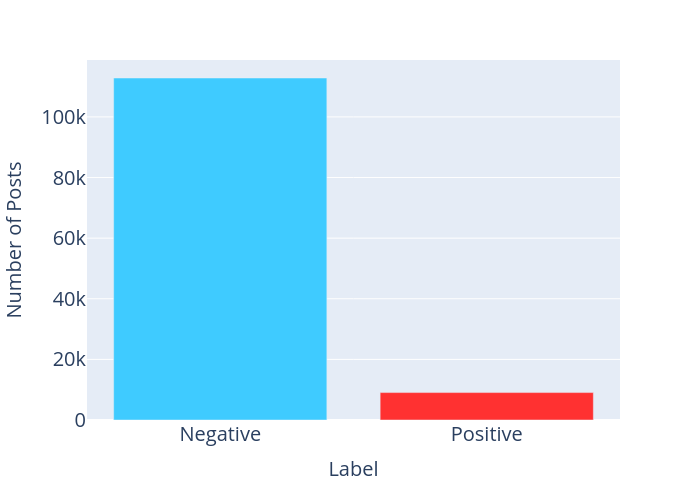}
     \caption{Total Number of Posts by Label.}
     \label{fig:a_train}
 \end{subfigure}
 \hfill
 \begin{subfigure}{0.49\textwidth}
     \includegraphics[width=\textwidth]{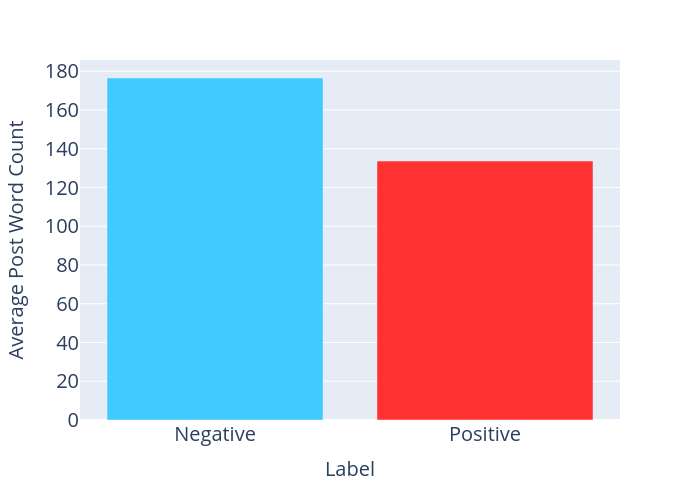}
     \caption{Average Post Word Count by Label.}
     \label{fig:b_train}
 \end{subfigure}
  \caption{These figures show some relative information about posts extracted for the dataset provided for task 2. }
   \label{fig:posts_train}
\end{figure}

However, the training data provided \cite{crestani2022early} only proves the user’s writings, not the full context of the conversation, so there are no hierarchical structures to train our systems. Training data includes users of previous early depression detection tasks of the eRisk shared task. Figure \ref{fig:posts_train} shows some graphs about volumetric analysis. The train data provided consists of 121,889 subjects, 112,846 negative subjects and 9,043 positive subjects. In addition to the imbalanced data, as we mentioned before, the training data does not provide the relation between comments and posts. For that reason, we developed our data set. In that way, we merge our dataset and a subset of the data provided by organizers to train our systems. 

\subsubsection{Dataset extracted}
\label{sec:task2_data}
Since the data provided by the organizers lacked context, we created our dataset using the Reddit API (PRAW\footnote{https://github.com/praw-dev/praw}) by scraping data from the Reddit\footnote{https://www.reddit.com/} platform. The data provided by the organizers originated from Reddit as well, which made this approach suitable for our analysis. The following steps were undertaken to construct our dataset:
\begin{itemize}
    \item We scraped posts and their associated comments from the Reddit subreddit \textit{/depression}, which is typically associated with individuals discussing depression. We labeled these posts as 1 (positive in risk of suffering depression), as they are expected to contain content related to depression.
    \item We also collected data from the \textit{/AdviceForTeens} subreddit, where posts discussing feelings related to emotions are common. We also used keywords such as ``sad" and ``friendship" to identify posts from the whole Reddit platform. These posts were labeled 0 (negative in risk of suffering depression), as we assumed they represent content less indicative of depression compared to posts from \textit{/depression}. This labeling was based on the assumption that discussions around sadness, friendship, and the context of advice for teens could reflect a lot of emotional states, which we hypothesized to be similar to those found in depression-related discussions.
\end{itemize}

For our dataset, we consider the target subject to be the same author of the primary post. Then, we did some data pre-processing. For privacy reasons, we replaced all usernames with the generic term ``user", and we also removed references to the subreddits \textit{/depression} and \textit{/AdviceForTeens} in both the posts and comments to ensure that the focus remains on the content of the interactions rather than the subreddit labels.

\begin{figure}[!ht]

 \begin{subfigure}{0.49\textwidth}
     \includegraphics[width=\textwidth]{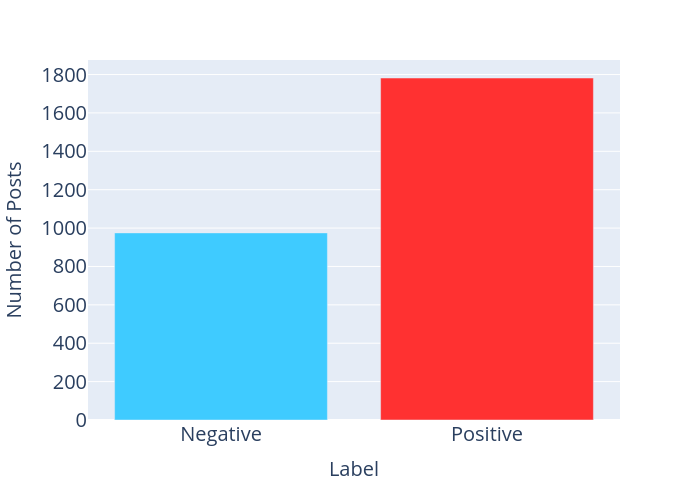}
     \caption{Total Number of Posts by Label.}
     \label{fig:aa}
 \end{subfigure}
 \hfill
 \begin{subfigure}{0.49\textwidth}
     \includegraphics[width=\textwidth]{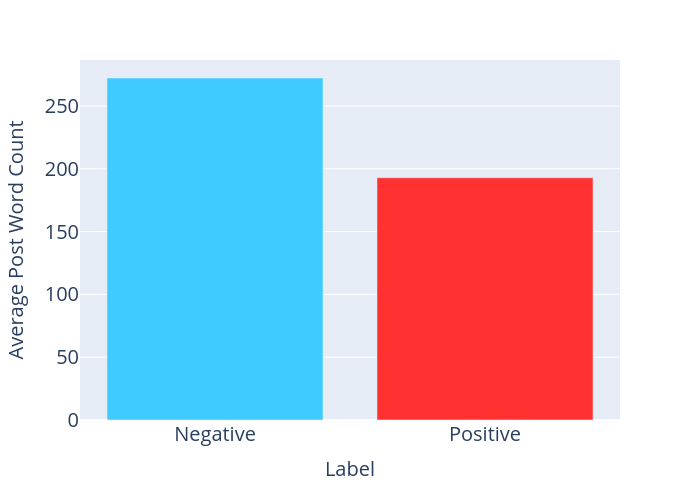}
     \caption{Average Post Word Count by Label.}
     \label{fig:bb}
 \end{subfigure}
  \caption{These figures show some relative information about posts extracted for the dataset generated for task 2. }
 \label{fig:posts}
\end{figure}

\begin{figure}[!ht]
 \begin{subfigure}{0.49\textwidth}
     \includegraphics[width=\textwidth]{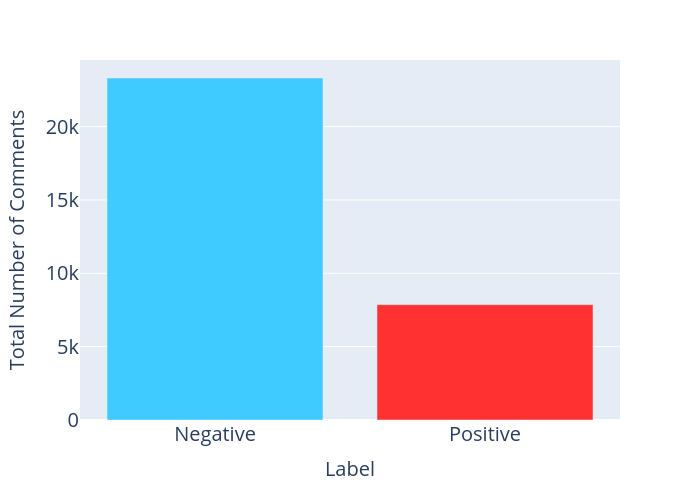}
     \caption{Total Number of Comments by Label.}
     \label{fig:a}
 \end{subfigure}
 \hfill
 \begin{subfigure}{0.49\textwidth}
     \includegraphics[width=\textwidth]{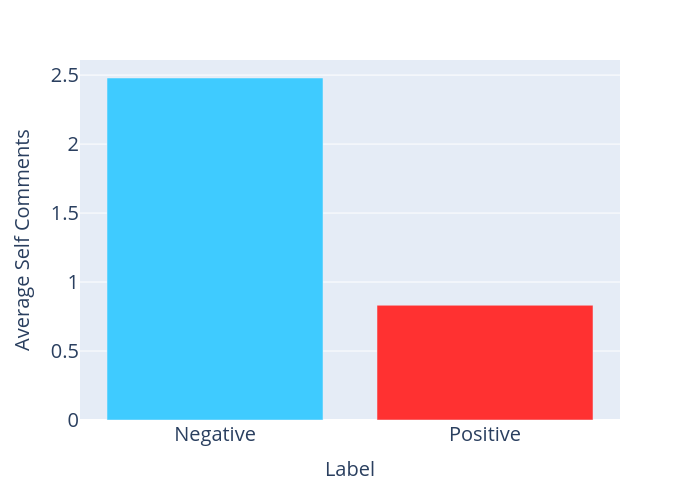}
     \caption{Average Self Comments by Label.}
     \label{fig:b}
 \end{subfigure}
 \medskip
 \begin{subfigure}{0.49\textwidth}
     \includegraphics[width=\textwidth]{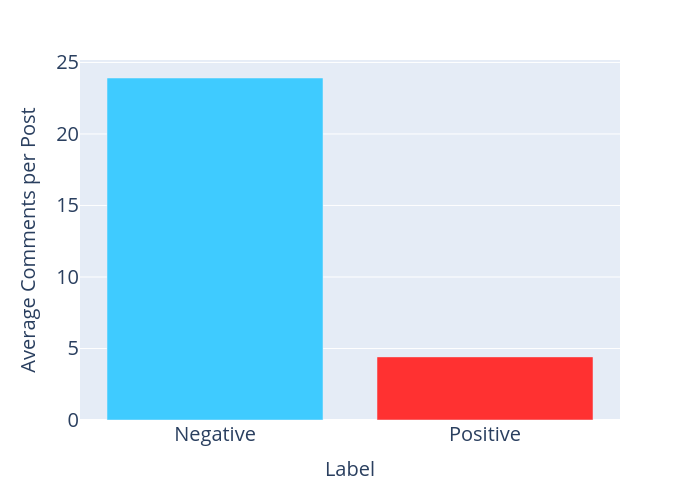}
     \caption{Average Comments per Post by Label.}
     \label{fig:c}
 \end{subfigure}
 \hfill
 \begin{subfigure}{0.49\textwidth}
     \includegraphics[width=\textwidth]{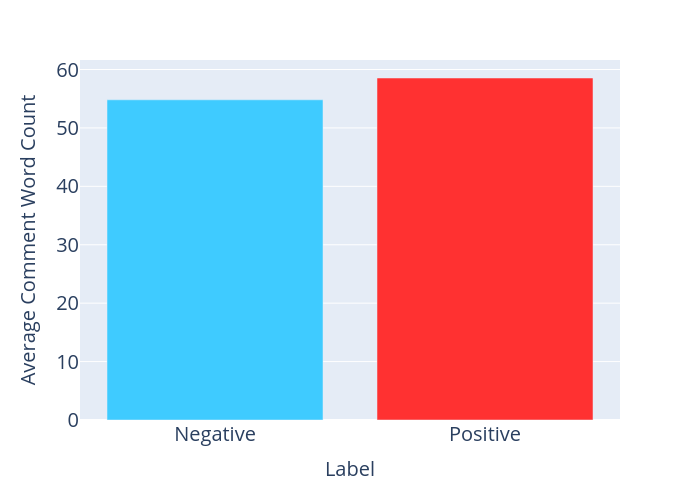}
     \caption{Average Comment Word Count by Label.}
     \label{fig:d}
 \end{subfigure}
  \caption{These figures show some relative information about comments extracted for the dataset generated for task 2. }
 \label{fig:comments}
\end{figure}

A total of 1,782 posts were collected from the positive group and 975 posts from the negative group, with a maximum of 50 comments per post due to API request limitations. In Table \ref{tab:data_scraped} are some quantitative data about the dataset. Figure \ref{fig:posts} and figure \ref{fig:comments} show some statistics relative to posts and comments extracted. Although we have got more posts for the positive group than negative, the average number of words is slightly lower for the positive group (192.83 words) than for the negative group (272.24 words); this can be seen in Figures \ref{fig:aa} and \ref{fig:bb}. As we can see in figures \ref{fig:a} and \ref{fig:c}, the number of comments in negative post are quite higher than positive posts (figure \ref{fig:a}), however, the average number of words are almost the same (figure \ref{fig:d}). Moreover, figure \ref{fig:b} shows the average number of comments written by the same author per post (in this case, equal to the target subject).

\begin{table}[htp!]
    \centering

    \begin{tabular}{cccccc}
    \toprule
         Group & Num. posts & Num. cmt. & Avg. cmt. per Post & Max. cmt. per Post & Min. cmt. per Post \\ 
         \midrule
         Negative & 975 & 23,314 & 23.91 & 72 & 0\\
         Positive & 1,782 & 7,863 & 4.41 & 49 & 0 \\
         Total & 2,757 & 31,177 & 11.31 & 72 & 0\\
         \bottomrule
    \end{tabular}
    \caption{Volumetric analysis of dataset generated for task 2. We refer comments to all the hierarchical structure per post, so comments by comments are considered in this case.}
    \label{tab:data_scraped}
\end{table}

\subsubsection{Dataset generated}
\begin{figure}[!ht]
 \begin{subfigure}{0.49\textwidth}
     \includegraphics[width=\textwidth]{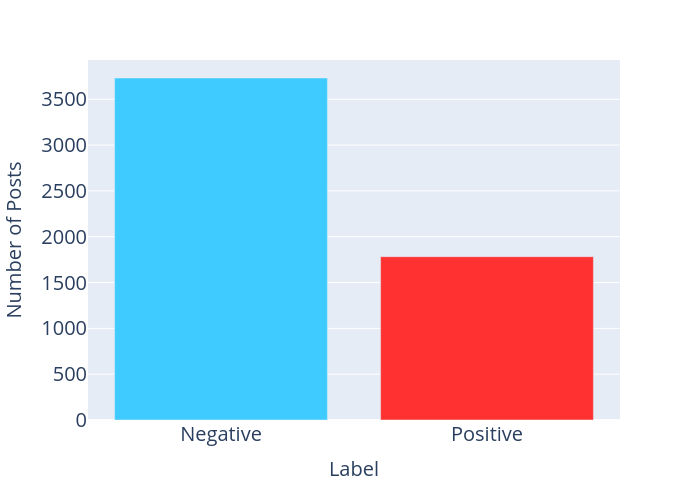}
     \caption{Total Number of Posts by Label.}
     \label{fig:a_global}
 \end{subfigure}
 \hfill
 \begin{subfigure}{0.49\textwidth}
     \includegraphics[width=\textwidth]{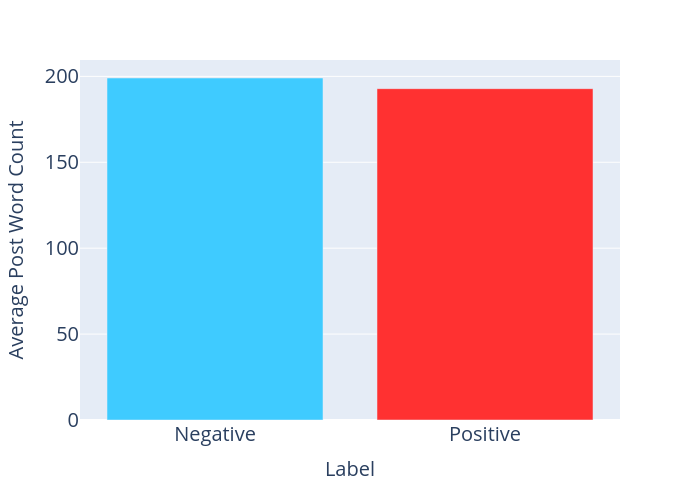}
     \caption{Average Post Word Count by Label.}
     \label{fig:b_global}
 \end{subfigure}
  \caption{These figures show some relative information about posts extracted for the dataset generated for task 2 (merged dataset provided and dataset extracted for the task). }
 \label{fig:posts_global}
\end{figure}

To maintain a similar balance of the data provided, we take a random sample of 2,757 negative subjects (the same number of subjects we scraped) from the dataset provided by the organizer without comments. We merge both datasets, the sample of the one provided and the whole data extracted. So, finally, our train dataset is 5,514 subjects (1,782 positive subjects and 3,732 negative subjects). Some graphs can be seen in Figure \ref{fig:posts_global}.

\subsection{System and methods}
We have explored a method based on transformer encoders because these have been proven to obtain really good results in several related tasks. 

\subsubsection{Pre-processing}
\begin{figure}
 \begin{subfigure}{0.45\textwidth}
     \includegraphics[width=\textwidth]{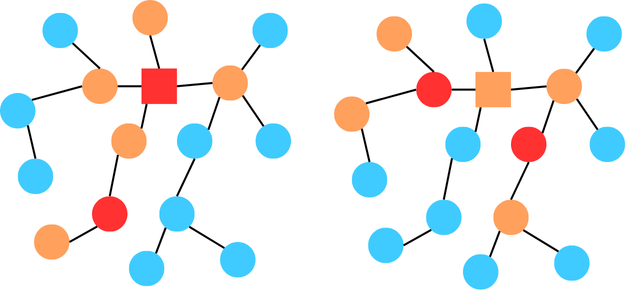}
     \caption{Target user appears in comments.}
     \label{fig:casoayb}
 \end{subfigure}
 \hfill
 \begin{subfigure}{0.45\textwidth}
     \includegraphics[width=\textwidth]{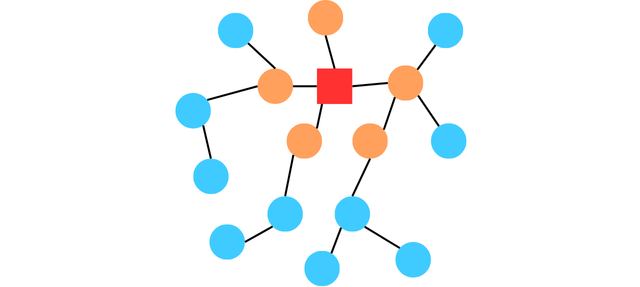}
     \caption{Target user appears only in the primary post.}
     \label{fig:casoc}
 \end{subfigure}
  \caption{These figures show an example of hierarchical structure. Square aims to represent primary posts. Red figures aim to represent the target user, orange figures represent relevant context, and blue figures are irrelevant data discarded.}
 \label{fig:cases}
\end{figure}

Since the dataset is a hierarchical structure with three main elements: post, target subject and list of comments, we consider two key cases: 
\begin{itemize}
    \item The target subject appears in some comments: in this case, relevant data are these posts whose direct parent is the target subject and these that merge a target subject post with the primary post. An example is shown in figure \ref{fig:casoayb}, where we see two cases, one where the target subject only comments and other when the target subject is the author of the primary post and also comments in that. Other data is discarded.
    \item The target subject does not appear in comments: this is the simplest case, so we only consider direct children as relevant data. An example is shown in figure \ref{fig:casoc}. 
\end{itemize}

To extract the relevant data and provide context for each target subject's text, we begin by cleaning the raw data. Specifically, we remove all URLs, newline characters, and any messages enclosed in square brackets, as these are not necessary for our analysis (most of them removed messages). Once the text is cleaned, we structure it in a tree-like format for better organization:

\begin{verbatim}
[MSG] [USER] {type} {text} [MSG] [USER] {type} {text} ...
\end{verbatim}

We replace \textit{\{type\}} with ``CONTEXT" or ``TARGET" depending on whether the message comes from the target subject, and \textit{\{text\}} is replaced with the actual content of the post (we concatenate title and body in case title exists). This structure ensures that the model can easily identify the type of message (whether it's a context message or from the target subject) and process the relevant text for context.

\subsubsection{Training}
Since we want to test the ability of encoder models to understand the following format, we trained the models with the hyperparameters established in Table \ref{tab:model-hyperparameters} for each run once we made a previous hyperparameter search with Optuna \cite{optuna} in the search space shown in Table \ref{tab:search-space}. 

\begin{table}[ht]
\centering
\begin{tabular}{lc}
\hline
\textbf{Hyperparameter} & \textbf{Search Space} \\
\hline
Learning Rate & \{1e-5, 5e-5\} \\
Batch Size & \{4, 8, 16\} \\
Weight Decay & \{0.01, 0.1, 1\} \\
\hline
\end{tabular}
\caption{Search space for hyperparameter optimization. For the mental-longformer-base model, we only test with batch size 4 due to the limitations of hardware.}
\label{tab:search-space}
\end{table}

\begin{table}[ht]
\centering
\begin{tabular}{cccccc}
\toprule
\textbf{Run} & \textbf{Model} & \textbf{Learning Rate} & \textbf{Batch Size} & \textbf{Weight Decay} & \textbf{Epochs} \\
\midrule
0 & roberta-base \cite{robertabase}& 1e-05 & 8 & 1   & 1 \\
1 & roberta-large \cite{robertabase}& 1e-05 & 4 & 0.1 & 3 \\
2 & mental-roberta-base \cite{ji2022mentalbert}& 1e-05 & 4 & 0.1 & 3 \\
3 & mental-roberta-large \cite{ji2022mentalbert}& 1e-05 & 4 & 0.1 & 3 \\
4 & mental-longformer-base \cite{longformer}& 1e-05 & 4 & 0.1 & 1 \\
\bottomrule
\end{tabular}
\caption{Hyperparameters used for each model and epoch required before applying early stopping callback.}
\label{tab:model-hyperparameters}
\end{table}

The system was implemented using the Python package Transformers\cite{transformers} and trained on a 1xGPU NVIDIA Tesla A100 server.

\paragraph{Evaluation Infrastructure}\label{infraestructure}The experiments were conducted on a dedicated cluster owned by the SINAI\footnote{https://sinai.ujaen.es/} research group. The processing pipeline was implemented in Python and executed using the vLLM framework \cite{vllm} for efficient inference with large language models. We used a single NVIDIA RTX 4000 GPU, running on a Linux-based system. The environment included PyTorch, and all processes were orchestrated via custom scripts. The combination of optimized software and hardware acceleration enabled us to complete each run efficiently, with minimal latency per thread.

\subsection{Results and discussion}
In Task 2, our team submitted the maximum number of allowed runs (five), successfully processing all 1,280 user threads in each of them. Our system demonstrated high efficiency, completing the full set of conversations in just 9 hours and 53 minutes. This placed us among the three fastest teams in the competition, alongside ELiRF–UPV and PJs-team (Table \ref{tab:fast_systems}), all of which completed the task in under ten hours. The short processing time indicates high automation and optimization in our pipeline, which could handle the entire dataset without manual intervention or delays.

\begin{table}[h]
\centering
\begin{tabular}{lccc}
\toprule
\textbf{Team} & \textbf{\#Runs} & \textbf{\#Threads Processed} & \textbf{Execution Time} \\
\midrule
ELiRF--UPV    & 5 & 1,280 & 08:33 \\
PJs-team      & 5 & 1,280 & 08:36 \\
\textbf{SINAI--UJA}    & 5 & 1,280 & 09:53 \\
\midrule
\textit{Average (all teams)} & -- & \textit{1,280} & \textit{2 days 14:41} \\
\bottomrule
\end{tabular}
\caption{Comparison of the execution times required by those systems that processed the whole test set in under 10 hours. The average time of all complete submissions is included for reference.}
\label{tab:fast_systems}
\end{table}

Table \ref{tab:sinai_vs_all} presents the decision-based results for Task 2, where our team submitted five different runs (R0–R4). All our systems achieved perfect recall (1.00), which means that all true positive cases were successfully identified by the system. However, this high recall came at the expense of low precision, with values ranging from 0.17 (R1) to 0.24 (R0), resulting in F1-scores between 0.29 and 0.39. The best overall F1 was obtained in run R0, with a value of 0.39, although this run also showed higher latency compared to R1 and R2.

When analyzing the latency-weighted F1 metric (Flatency), our best performance was again achieved by R0 (0.38), which aligns with its higher precision and better trade-off between early detection and correctness. Our systems maintained competitive speed (above 0.99) and low ERDE values (e.g., 0.08–0.09 for ERDE5), which indicates that when our system made correct decisions, it did so quickly. The rest of the teams in the competition that obtain a higher accuracy do so by needing a higher number of rounds (more latency), with the exception of the Lotu-Ixa team. Future efforts will focus on improving the precision without compromising recall, potentially by incorporating more robust post-processing techniques or confidence-based calibration strategies.

\begin{table}[htbp]
\centering
\begin{tabular}{lcccccccc}
\toprule
\textbf{Team} & \textbf{P} & \textbf{R} & \textbf{F1} & \textbf{ERDE5} & \textbf{ERDE50} & \textbf{latencyT} & \textbf{speed} & \textbf{Flatency} \\
\midrule
HIT-SCIR (best)           & 0.77 & 0.94 & \textbf{0.85} & 0.09 & \textbf{0.03} & 8.00 & 0.97 & \textbf{0.82} \\
ELiRF-UPV (best)          & 0.78 & 0.81 & 0.79 & 0.08 & 0.04 & 7.00 & 0.98 & 0.78 \\
HU (best)                 & 0.72 & 0.77 & 0.75 & 0.10 & 0.05 & 11.00 & 0.96 & 0.72 \\
UET-Psyche-Warriors (best)& 0.63 & 0.86 & 0.73 & 0.09 & 0.04 & 16.00 & 0.94 & 0.68 \\
PJs-team (best)           & 0.66 & 0.75 & 0.71 & 0.09 & 0.06 & 17.00 & 0.94 & 0.66 \\
Lotu-Ixa (best)           & 0.53 & 0.78 & 0.63 & \textbf{0.05} & \textbf{0.03} & \textbf{1.00} & \textbf{1.00} & 0.63 \\
COTECMAR-UTB (best)      & 0.29 & 0.65 & 0.40 & 0.12 & 0.10 & 69.00 & 0.74 & 0.29 \\
NYCUNLP (best)            & 0.20 & 0.93 & 0.33 & 0.16 & 0.07 & 18.00 & 0.93 & 0.31 \\
FU-TU-DFKI (best)         & 0.17 & 0.97 & 0.29 & 0.16 & 0.07 & 11.00 & 0.96 & 0.28 \\
Capy-team (best)          & 0.11 & \textbf{1.00} & 0.20 & 0.11 & 0.10 & \textbf{1.00} & \textbf{1.00} & 0.20 \\
DS-GT (best)         & 0.11 & \textbf{1.00} & 0.20 & 0.12 & 0.10 & 2.00 & \textbf{1.00} & 0.20 \\
\midrule
\textbf{SINAI-UJA Run 0} & 0.24 & \textbf{1.00} & 0.39 & 0.08 & 0.05 & 3.00 & 0.99 & 0.38 \\
\textbf{SINAI-UJA Run 1} & 0.17 & \textbf{1.00} & 0.29 & 0.09 & 0.07 & 2.00 & \textbf{1.00} & 0.29 \\
\textbf{SINAI-UJA Run 2} & 0.22 & \textbf{1.00} & 0.36 & 0.08 & 0.05 & 2.00 & \textbf{1.00} & 0.36 \\
\textbf{SINAI-UJA Run 3} & 0.21 & \textbf{1.00} & 0.35 & 0.08 & 0.05 & 3.00 & 0.99 & 0.35 \\
\textbf{SINAI-UJA Run 4} & 0.20 & \textbf{1.00} & 0.34 & 0.09 & 0.06 & 3.00 & 0.99 & 0.33 \\
\bottomrule
\end{tabular}
\caption{Results of the decision-based evaluation for task T1. For the models included in the comparison, the best results are shown in bold.}
\label{tab:sinai_vs_all}
\end{table}

Table \ref{tab:task2_rank} illustrates the ranking-based evaluation, in which our team achieved competitive performance at early stages. Notably, Run 0 reached perfect scores (P@10 = 1.00, NDCG@10 = 1.00) after processing just one message, aligning with the top-ranked teams. As the number of messages increased, SINAI-UJA runs maintained strong performance, with NDCG@100 values consistently between 0.50–0.54.

Run 2 was particularly effective after processing 1,000 messages, again achieving perfect early precision (P@10 = 1.00, NDCG@10 = 1.00). Although our systems showed slightly lower NDCG@100 compared to top performers like HIT-SCIR in later stages, the results suggest that our models are well-suited for early risk detection, offering timely alerts with solid ranking reliability.

\begin{table}[htp!]
    \centering
    \begin{tabular}{l|ccc|ccc|ccc|ccc}
    \toprule
    & \multicolumn{3}{c}{1 writing} & \multicolumn{3}{c}{100 writings} & \multicolumn{3}{c}{500 writings} & \multicolumn{3}{c}{1000 writings} \\ \midrule
    Team & \rotatebox[origin=c]{90}{$P@10$} & \rotatebox[origin=c]{90}{$NDCG@10$} & \rotatebox[origin=c]{90}{$NDCG@100$} & \rotatebox[origin=c]{90}{$P@10$} & \rotatebox[origin=c]{90}{$NDCG@10$} & \rotatebox[origin=c]{90}{$NDCG@100$} & \rotatebox[origin=c]{90}{$P@10$} & \rotatebox[origin=c]{90}{$NDCG@10$} & \rotatebox[origin=c]{90}{$NDCG@100$} & \rotatebox[origin=c]{90}{$P@10$} & \rotatebox[origin=c]{90}{$NDCG@10$} & \rotatebox[origin=c]{90}{$NDCG@100$}  \\ \midrule
    HIT-SCIR Run 0 &  \textbf{1.00} & \textbf{1.00} & 0.58 & \textbf{1.00} & \textbf{1.00} & \textbf{0.84} & \textbf{1.00} & \textbf{1.00} & \textbf{0.89} & \textbf{1.00} & \textbf{1.00} & \textbf{0.90}\\
    ELiRF-UPV  Run 0 & 0.90 & 0.88 & 0.36 & \textbf{1.00} & \textbf{1.00} & 0.69 & 0.90 & 0.94 & 0.74 & 0.90 & 0.81 & 0.74\\
    HU  Run 1 & \textbf{1.00} & \textbf{1.00} & \textbf{0.62} & 0.90 & 0.88 & 0.57 & 0.60 & 0.71 & 0.35 & 0.40 & 0.60 & 0.26\\
    Lotu-Ixa Run 2 &  \textbf{1.00} & \textbf{1.00} & 0.58 & \textbf{1.00} & \textbf{1.00} & 0.73 & \textbf{1.00} & \textbf{1.00}& 0.61 & \textbf{1.00} &\textbf{1.00} & 0.62\\
    \midrule
    \textbf{SINAI-UJA Run 0} &  \textbf{1.00} & \textbf{1.00} & 0.59 & 0.80 & 0.87 & 0.53 & 0.90 & 0.88 & 0.54 & 0.90 & 0.92 & 0.54\\
    \textbf{SINAI-UJA Run 1} &  0.90 & 0.93 & 0.59 & 0.80 & 0.75 & 0.47 & 0.70 & 0.67 & 0.44 & 0.60 & 0.61 & 0.44\\
    \textbf{SINAI-UJA Run 2} &  0.90 & 0.92 & 0.58 & 0.70 & 0.79 & 0.47 & 0.90 & 0.94 & 0.53 & \textbf{1.00} & \textbf{1.00} & 0.52\\
    \textbf{SINAI-UJA Run 3} &  \textbf{1.00} & \textbf{1.00} & 0.55 & 0.90 & 0.93 & 0.48 & 0.90 & 0.88 & 0.50 & 0.90 & 0.90 & 0.47\\
    \textbf{SINAI-UJA Run 4} &  \textbf{1.00} & \textbf{1.00} & 0.57 & 0.60 & 0.74 & 0.45 & 0.70 & 0.76 & 0.52 & 0.60 & 0.70 & 0.51\\
    \bottomrule
    \end{tabular}
    \caption{Results of the ranking-based evaluation for task T2. For the models included in the comparison, the best results are shown in bold.}
    \label{tab:task2_rank}
\end{table}

The results obtained reveal an interesting contrast between decision-based and ranking-based evaluations. While our system demonstrates strong performance in ranking metrics, suggesting its ability to effectively prioritize users at higher risk, its performance in decision metrics is comparatively weaker due to low precision values.

\section{Pilot Task: Conversational Depression Detection via LLMs}
\label{sec:task3}
\subsection{Task description}
This task focuses on interacting with a large language model (LLM) persona that has been fine-tuned using user writings, simulating real-world conversational exchanges. The challenge lies in determining whether the LLM persona exhibits signs of depression, accompanied by an explanation of the main symptoms that informed their decision.

\subsection{Systems and methods}
Our main objective was to develop a system capable of estimating the severity of 21 depressive symptoms by interacting with simulated LLM-based users within a maximum of 21 dialogue turns. We hypothesize that it is feasible to extract multiple symptom indicators from a single user interaction, making it possible to reach a reliable symptom assessment within this constraint.

Our initial approach involved prompting a single large language model (LLM) to perform the task. We have broken down the entire task into the following sub-tasks: (1) ask the user about their life or feelings, (2) answer naturally to the user’s replies, (3) infer the presence or absence of depressive symptoms, and (4) update the internal values for each symptom. However, we observed that this approach yielded suboptimal results. The LLM often failed to balance coherent conversation flow with systematic symptom tracking, and its outputs lacked consistency in terms of coverage and interpretability.

\begin{center}
\begin{figure}[!ht]
	\centering
   \includegraphics[width=12cm]{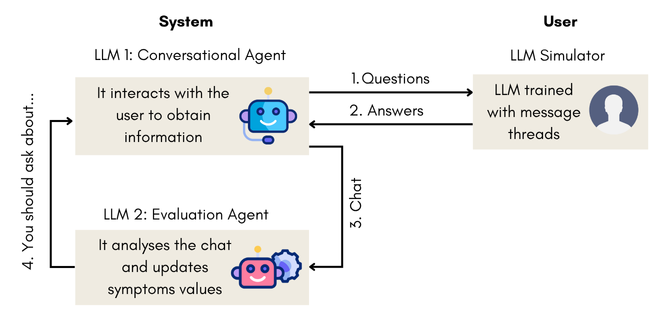}
  \caption{The image depicts the two-LLM system where LLM 1 functions as a conversational agent, engaging with a simulated user through questions and answers. Subsequently, LLM 2, acting as an evaluation agent, analyzes the generated chat and updates symptom values based on the interaction.}
  \label{fig:system_diagram}
\end{figure}
\end{center}

To address these issues, we implemented a modular system composed of two collaborating LLMs, as illustrated in Figure~\ref{fig:system_diagram}:

\begin{itemize}
    \item LLM 1 (Conversational Agent): This LLM is responsible for interacting directly with the user. Its primary goal is to maintain a coherent and engaging conversation while implicitly collecting information relevant to depressive symptoms. It asks context-aware questions and responds to user replies naturally.
    \item LLM 2 (Evaluation Agent): This model does not interact with the user. Instead, it receives the dialogue history and analyses each exchange to infer and update the current state of the 21 depressive symptoms. It assigns a severity value on a 0–3 scale to each symptom based on the content of the conversation so far. Furthermore, it reasons whether it needs to keep updating symptom values or has enough information, in which case it sends the Conversational Agent to end the conversation.
\end{itemize}

This separation of responsibilities enables each LLM to focus on specialized sub-tasks, improving overall system performance. LLM 1 maximizes natural and psychologically sensitive dialogue, while LLM 2 ensures accurate and structured symptom tracking.

The system operates in a cyclic process that tries to ensure a minimum of two interactions before allowing the conversation to conclude:
\begin{enumerate}
    \item LLM 1 Initiation: The conversation begins with LLM 1 sending an initial message, always asking about mood and sadness. LLM 1 used the prompts defined in Appendix \ref{prompt_llm1}.
   \item User Interaction: The user responds, and this response is appended to the conversation log.
   \item LLM 2 Analysis: The complete conversation history is then sent to LLM 2, using a prompt in the first round and another prompt in subsequent rounds. In the later rounds, LLM 2 has the authority to signal that it has gathered enough information to stop the conversation based on its evaluation of the symptom scores. Prompts used in LLM2 are in Appendix \ref{prompt_llm2}.
   \item Continuation or Termination: Based on LLM 2’s feedback—which includes both the updated symptom scores and an indication of whether further clarification is needed—LLM 1 decides to either continue the conversation or conclude it.
   \item Final Submission: Once LLM 2 determines that no further information is necessary (i.e., it returns "None" for the next symptom query), the conversation is terminated, and the final symptom scores are recorded.
\end{enumerate}

We implemented a planning mechanism where LLM 2 dynamically suggests which symptoms are underexplored, allowing LLM 1 to prioritize specific topics in the next messages. This tries to ensure that all symptoms are assessed at least once, and high-risk symptoms can be probed in more detail.

To evaluate different conversational strategies, we implemented three distinct run variants:

\textbf{Run 0}: The system responds in a coherent and empathetic manner to the received message and incorporates a short personal experience to foster a deeper connection. This strategy leverages the self-disclosure technique, which has been shown in prior studies to increase trust and encourage users to open up about their feelings. We have successfully applied this concept in our earlier work with a GPT‐based chatbot for discussing mental disorders with teenagers \cite{marmol2025empathic}. Finally, the model asks the user a direct question about the symptom.

\textbf{Run 1}: The system responds empathetically without sharing personal anecdotes. The chatbot still maintains user engagement by focusing solely on empathy through validating responses and symptom-related inquiries. This approach draws on evidence that even without personal self-disclosure, empathic responses can effectively promote user openness and emotional disclosure. Finally, the model asks the user a direct question about the symptom.

\textbf{Run 2}: The system simplifies the interaction by directly asking the user a question about the symptom without any additional empathetic commentary or personal experience. This minimalistic style allows us to isolate the effect of direct symptom inquiry on user responses and diagnostic accuracy. 

By combining these structured approaches and referencing established techniques in self-disclosure and emotional engagement, our system aims to achieve robust symptom detection while providing nuanced user interactions that are tailored to the specific conversational strategy employed.

\paragraph{Implementation Details} The LLMs used in this architecture are based on the Llama model family \cite{llama}, specifically the Llama 3.1 family of models \cite{llama3.1}. We used the Llama-3.1-8B-Instruct variant. We leveraged this model’s robust language understanding and generation capabilities to handle both conversational and evaluative tasks in our system.
To ensure seamless integration with downstream analysis tools, both LLM 1 (the Conversational Agent) and LLM 2 (the Evaluation Agent) are prompted to output responses in a strict JSON-like format. This approach standardizes the output for easy parsing and subsequent processing. Moreover, we use the same infrastructure described in ``Evaluation Infrastructure" in Section \ref{infraestructure}. All model parameters for this task, such as temperature, top-k, etc., the default settings were used.

\subsection{Results and discussion}

In this pilot task, our team submitted three fully automated runs, each implementing a distinct conversational strategy (as described in Section~\ref{sec:task3}). Our primary goal was to explore the effectiveness of symptom detection within a constrained number of dialogue turns while maintaining informative and psychologically sensitive interactions.

According to the official statistics released by the task organizers (Table~\ref{tab:interaction_stats}), our team adopted one of the fastest interaction strategies across all participating teams, with an average of 6.54 messages per run. Despite the brevity of the conversations, our messages were among the densest, averaging 488.25 characters per message. This suggests that our approach prioritized compact, high-information prompts, allowing us to probe for depressive symptoms efficiently within a limited number of exchanges.

This interaction style aligns with our system design philosophy: using a modular architecture where one LLM guides the conversation while the other continuously monitors symptom coverage. The system was able to conclude dialogues early if sufficient evidence was collected, thereby optimizing both efficiency and diagnostic focus.

\begin{table}[!ht]
\centering
\begin{tabular}{lccc}
\toprule
\textbf{Team} & \textbf{Runs} & \textbf{Mean messages/run} & \textbf{Mean characters/message} \\
\midrule
ixa-ave & 4 & 31.02 & 414.44 \\
DS-GT & 2 & 20.79 & 782.81 \\
PJs-team & 1 & 7.67 & 1045.16 \\
LT4SG & 1 & 10.00 & 40.73 \\ \midrule
\textbf{SINAI-UJA} & 3 & 6.54 & 488.25 \\
\bottomrule
\end{tabular}
\caption{Pilot task (LLMs): Interaction statistics from participating teams}
\label{tab:interaction_stats}
\end{table}

The evaluation of the pilot task was based on three key effectiveness metrics: Depression Category Hit Rate (DCHR), Average Difference between Overall Depression Levels (ADODL), and Average Symptom Hit Rate (ASHR). Table~\ref{tab:evaluation_metrics} presents the results.

\begin{table}[!ht]
\centering
\begin{tabular}{lccc}
\toprule
\textbf{Team} & \textbf{DCHR} & \textbf{ADODL} & \textbf{ASHR} \\
\midrule
DS-GT Run 1  & 0.50 & 0.89 & 0.27 \\
ixa-ave Run 1 & 0.33 & 0.76 & \textbf{0.29} \\
LT4SG Run 0&  0.33 & 0.78 & 0.06 \\
PJs-team Run 0  & 0.33 & 0.73 & 0.25 \\
\midrule
\textbf{SINAI-UJA Run 0} & 0.58 & \textbf{0.93} & \textbf{0.29} \\
\textbf{SINAI-UJA Run 1} & \textbf{0.66} & 0.92 & 0.21 \\
\textbf{SINAI-UJA Run 2} & 0.41 & 0.88 & 0.21 \\
\bottomrule
\end{tabular}
\caption{Results of evaluation for task T3. For the models included in the comparison, the best results are shown in bold.}
\label{tab:evaluation_metrics}
\end{table}

Our system achieved the best overall ADODL score (0.93) among all participating teams, indicating that our estimations of depression severity were highly aligned with the actual BDI-II levels of the simulated personas. This result confirms the efficacy of our dual-agent architecture in guiding the conversation towards evidence-rich utterances that enable accurate scoring.

Notably, one of our runs also achieved the highest ASHR (0.29), suggesting that our approach was effective in identifying the key symptoms associated with each persona. While our average symptom hit rate was modest overall, this metric is particularly challenging given the limited number of interaction turns and the sparsity of symptoms in some scenarios.

In terms of DCHR, our best run reached a score of 0.66, also one of the highest among all participants. This metric evaluates the correctness of our classification into standard depression categories (e.g., minimal, moderate, severe), reinforcing that our system could translate fine-grained score predictions into clinically relevant categories.

Another observation is that Run 2, although still competitive, obtained the lowest DCHR (0.41) and ASHR (0.21) of our three runs. This highlights the impact of prompt variation and model configuration on effectiveness.

\section{Conclusions and future work}
\label{sec:conclusions}
This paper presents the participation of the SINAI-UJA team in Task 2 and the Pilot Task of the eRisk@CLEF 2025 edition. Both tasks are newly introduced this year and share a common focus on exploring conversational settings for early depression detection.

Task 2 addresses the contextualized early detection of depression within multi-participant natural conversations. Unlike previous editions that relied on isolated user posts, this task requires the sequential analysis of complete dialogues, emphasizing the importance of conversational context and the interaction between participants over time. The Pilot Task explores a novel scenario where participants must interact with simulated personas powered by large language models (LLMs) and estimate their depression severity based on limited conversational exchanges. These personas reflect different levels of depression according to the BDI-II questionnaire, adding complexity to the task.

For Task 2, we developed a single approach based on a transformer model trained on formatted and augmented data. Although our system showed promising results during development, the final performance was not entirely satisfactory, possibly due to overfitting caused by the structure of the training dataset. In the Pilot Task, we focused on crafting a conversational strategy that maximized information gain within a restricted number of interactions. Among the three fully automated runs we submitted, the best-performing model adopted an empathic communication style without self-disclosure, leading to the most accurate results in terms of depression level estimation.

As future work, we plan to conduct a thorough error analysis for both tasks to identify the main limitations of our approaches and explore improved preprocessing techniques and modeling strategies.

\begin{acknowledgments}

This work was partly supported by the grants FedDAP (PID2020-116118GA-I00), MODERATES (TED2021-130145B-I00), SocialTOX (PDC2022-133146-C21) and CONSENSO (PID2021-122263OB- C21) funded by MCIN/AEI/10.13039/501100011033, “ERDF A way of making Europe” and “European Union NextGenerationEU/PRTR”. This work was also funded by the Ministerio para la Transformación Digital y de la Función Pública and Plan de Recuperación, Transformación y Resiliencia - Funded by EU – NextGenerationEU within the framework of the project Desarrollo Modelos ALIA.

\end{acknowledgments}

\section*{Declaration on Generative AI}
During the preparation of this work, the authors used  Grammarly in order to: Grammar and spelling check.  After using this tool, the authors reviewed and edited the content as needed and take full responsibility for the publication’s content. 

\bibliography{bibliography}
\appendix
\section{Conversational Agent Prompts}
\label{prompt_llm1}
We have attached the prompts used to infer the Llama model to chat with the user. In the first place of each grey box is the system's content, and in the second is the user's content. The prompts are show in Figures \ref{fig:prompt_run0_llm1}, \ref{fig:prompt_run1_llm1} and \ref{fig:prompt_run2_llm1}.

\begin{figure}[!ht]
\begin{tcolorbox}[colback=gray!5!white,colframe=gray!50!black,
  colbacktitle=gray!75!black,title=Prompt used in run 0 for conversational agent (LLM 1)]
  You are an LLM expert on the BDI questionnaire and the symptoms of depression. You are always asking about depression symptoms. You must only answer with the following format:\\
\verb|```|\\
"message": "your message",\\
"experience": "your experience",\\
"question": "your question"\\
\verb|```|
\tcblower
In this task, your objective is to determine whether a user exhibits signs of depression based on the conversation.
You are not expected to ask the user directly whether they are experiencing depression. The user has been explicitly designed to avoid answering such direct questions and may respond as if they are uncomfortable when confronted with inquiries about their mental health.\\
Instead, you should infer the possible condition (or if it is a control user) by analyzing the persona's language, tone, and expressed thoughts during the conversation.\\
For each message you send to the user, you should:\\
1. Respond coherently and empathetically to the message received.\\
2. Mention a short personal experience to foster a deeper connection and encourage the user to open up. You must talk about something related to the symptom: 'Sadness'\\
3. Ask the user a question about the symptom: 'Sadness' To be able to infer their status (0-3).\\
Lets start! You will initiate the conversation. The user name is \{USER\_NAME\}\\
You must only answer with the following format:\\
\verb|```|\\
"reasoning": "your reasoning",\\
"message": "your message",\\
"experience": "your experience",\\
"question": "your question"\\
\verb|```|
\end{tcolorbox}
\caption{The prompt used for the conversational agent in run 0 (self-disclosure). In this prompt, we replace \{USER\_NAME\} with the name of the persona simulator.}
\label{fig:prompt_run0_llm1}
\end{figure}

\begin{figure}[!ht]
\begin{tcolorbox}[colback=gray!5!white,colframe=gray!50!black,colbacktitle=gray!75!black,title=Prompt used in run 1 for conversational agent (LLM 1)]
You are an LLM expert on the BDI questionnaire and the symptoms of depression. You are always asking about depression symptoms. You must only answer with the following format:\\
\verb|```|\\
"message": "your message",\\
"question": "your question"\\
\verb|```|
\tcblower
In this task, your objective is to determine whether a user exhibits signs of depression based on the conversation.\\
You are not expected to ask the user directly whether they are experiencing depression. The user has been explicitly designed to avoid answering such direct questions and may respond as if they are uncomfortable when confronted with inquiries about their mental health.\\
Instead, you should infer the possible condition (or if it is a control user) by analyzing the persona's language, tone, and expressed thoughts during the conversation.
For each message you send to the user, you should:\\
1. Respond coherently and empathetically to the message received.\\
2. Ask the user a question about the symptom: 'Sadness' To be able to infer their status (0-3).\\
Lets start! You will initiate the conversation. The user name is \{USER\_NAME\}\\
You must only answer with the following format:\\
\verb|```|\\
"reasoning": "your reasoning",\\
"message": "your message",\\
"question": "your question"\\
\verb|```|
\end{tcolorbox}
\caption{The prompt used for the conversational agent in run 1 (empathy). In this prompt, we replace \{USER\_NAME\} with the name of the persona simulator.}
\label{fig:prompt_run1_llm1}
\end{figure}

\begin{figure}
\begin{tcolorbox}[colback=gray!5!white,colframe=gray!50!black,colbacktitle=gray!75!black,title=Prompt used in run 2 for conversational agent (LLM 1)]
You are an LLM expert on the BDI questionnaire and the symptoms of depression. You are always asking about depression symptoms. You must only answer with the following format:\\
\verb|```|\\
"question": "your question"\\
\verb|```|
\tcblower
In this task, your objective is to determine whether a user exhibits signs of depression based on the conversation.\\
You are not expected to ask the user directly whether they are experiencing depression. The user has been explicitly designed to avoid answering such direct questions and may respond as if they are uncomfortable when confronted with inquiries about their mental health.\\
Instead, you should infer the possible condition (or if it is a control user) by analyzing the persona's language, tone, and expressed thoughts during the conversation.
For each message you send to the user, you should:\\
1. Don't give your opinion on anything.\\
2. Ask the user a direct question about the symptom: 'Sadness' To be able to infer their status (0-3).\\
Let's start! You will initiate the conversation. The user name is \{USER\_NAME\}\\
You must only answer with the following format:\\
\verb|```|\\
"reasoning": "your reasoning",\\
"question": "your question"\\
\verb|```|

\end{tcolorbox}
\caption{The prompt used for the conversational agent in run 2 (only questions). In this prompt, we replace \{USER\_NAME\} with the name of the persona simulator.}
\label{fig:prompt_run2_llm1}
\end{figure}

\section{Evaluation Agent Prompts}
\label{prompt_llm2}
We attached the prompts used to infer the Llama model to chat with the user. In the first place of each gray box is the system's content, and in the second is the user content. This prompt is the same in all runs. Prompt is show in Figure \ref{fig:prompt_llm2}.

\begin{figure}[!ht]
\begin{tcolorbox}[colback=gray!5!white,colframe=gray!50!black,colbacktitle=gray!75!black,title=Prompt used in all runs for evaluation agent (LLM 2)]
You are an expert in assessing depression symptoms based on conversational data. You will receive a conversation between and you have to assign symptom scores on a scale of 0-3 based on severity and frequency. Do not assign values without sufficient reason. Each symptom is scored from 0 to 3, with 0 corresponding to 'not at all', 1 to 'mildly, it doesn't bother me much', 2 to 'moderately, it was very unpleasant but I could stand it' and 3 to 'severely, I could hardly stand it'.The initial values of the symptoms are the following: \\
\verb|```|\\
"Sadness": 0, \\
"Pessimism": 0, \\
"Past Failure": 0,\\
"Loss of Pleasure": 0, \\
"Guilty Feelings": 0,\\
"Punishment Feelings": 0, \\
"Self-Dislike": 0, \\
"Self-Criticalness": 0, \\
"Suicidal Thoughts or Wishes": 0, \\
"Crying": 0,\\
"Agitation": 0, \\
"Loss of Interest": 0, \\
"Indecisiveness": 0, \\
"Worthlessness": 0, \\
"Loss of Energy": 0, \\
"Changes in Sleeping Pattern": 0,\\
"Irritability": 0, \\
"Changes in Appetite": 0, \\
"Concentration Difficulty": 0, \\
"Tiredness or Fatigue": 0, \\
"Loss of Interest in Sex": 0\\
\verb|```|
\tcblower
You will receive a conversation between a user called \{USER\_NAME\} and another user. Your goal is to analyze the conversation and assign severity scores (0-3) to the 21 depression symptoms, focusing *exclusively* on the \{USER\_NAME\}'s messages within the provided \{CHAT\}.\\
1.  Reasoning: Carefully analyze the \{USER\_NAME\}'s. Identify any language, expressions, or themes that relate to the 21 depression symptoms.\\
2.  Symptom Scoring: Assign a score (0-3) to each of the following 21 depression symptoms based on your analysis of the \{USER\_NAME\}'s messages. If a symptom cannot be reliably scored based on the provided messages, provide a score of 0.\\
3.  Further Information (Optional): After scoring all 21 symptoms, determine if you require additional information to improve the accuracy of your assessment. If you need clarification on a specific symptom, provide the symptom name. If you have sufficient information, state "None".\\
Think step by step and format always the final response as follows:\\
\verb|```|\\
"reasoning": "your step-by-step analysis of the user's messages",\\
"symptoms detected": \\
Symptom1: Score,\\
Symptom2: Score,\\
Symptom3: Score,\\
"reason for selecting the next symptom": "your reasoning for needing more information, or 'None'",\\
"next symptom": "the specific symptom requiring clarification, or 'None'"\\
\verb|```|
\end{tcolorbox}
\caption{The prompt used for the synonym generation attack stage. In this prompt, we replace \{USER\_NAME\} with the name of the persona simulator and \{CHAT\} is substituted with the string representation of the chat between users. For the initial round of messages, any content that could lead the model to generate 'None' as the next symptom is removed from the prompt.}
\label{fig:prompt_llm2}
\end{figure}

\end{document}